\pdfoutput=1

\documentclass[11pt,a4paper]{article}
\usepackage[hyperref]{paper_arxiv}
\usepackage{times}
\usepackage{latexsym}

\usepackage{microtype}
\usepackage{inconsolata}

\usepackage{multirow}
\usepackage{graphicx}
\usepackage{url}
\usepackage{amsmath}
\usepackage{amssymb}
\usepackage{cprotect}
\usepackage{enumitem}

\aclfinalcopy 

\title{Controlling keywords and their positions in text generation}

\author{%
	\textbf{Yuichi Sasazawa, Terufumi Morishita, Hiroaki Ozaki}, \\
	\textbf{Osamu Imaichi, Yasuhiro Sogawa} \\
	Hitachi, Ltd. Research and Development Group \\ 
	\texttt{\{yuichi.sasazawa.bj, terufumi.morishita.wp, hiroaki.ozaki.yu,}\\
	\texttt{osamu.imaichi.xc, yasuhiro.sogawa.tp\}@hitachi.com}
}

\begin{document}
	
	\maketitle
	\begin{abstract}
		One of the challenges in text generation is to control text generation as intended by the user. Previous studies proposed specifying the keywords that should be included in the generated text. However, this approach is insufficient to generate text that reflect the user's intent. For example, placing an important keyword at the beginning of the text would help attract the reader's attention; however, existing methods do not enable such flexible control. In this paper, we tackle a novel task of controlling not only keywords but also the position of each keyword in the text generation. To this end, we propose a task-independent method that uses special tokens to control the relative position of keywords. Experimental results on summarization and story generation tasks show that the proposed method can control keywords and their positions. The experimental results also demonstrate that controlling the keyword positions can generate summary texts that are closer to the user's intent than baseline.

	\end{abstract}
	
	\section{Introduction}
	One of the challenges in text generation is to generate text that is consistent with the user's intent. Many methods for specifying the keywords that should be included in the generated text to reflect the user's intent have been proposed. As for summarization, by providing the model with keywords that should be included in the summary, it is possible to generate summaries that focus on specific parts of the document~\cite{fan-etal-2018-controllable, he-etal-2022-ctrlsum, dou-etal-2021-gsum}. As for story generation, keywords are used to control the narrative storyline~\cite{DBLP:journals/corr/JainAMSLS17, fan-etal-2019-strategies, 10.1609/aaai.v33i01.33017378}. As for other tasks, such as e-commerce generation, review generation, and question generation, keywords are also used to control text generation~\cite{chan-etal-2019-stick, 10.1145/3442381.3449838, ni-mcauley-2018-personalized, DBLP:journals/corr/abs-2112-01012, 10.1145/3442381.3449876}. In addition, more-advanced methods that specify the order of keywords to be included in the generated text to control the rough storyline have been proposed~\cite{su-etal-2021-plan-generate, 10.1145/3442381.3449838}.
	
	The above-described methods, however, cannot generate texts that reflect more fine-grained intentions. Specifically, the user may want to reflect the intended importance of each keyword in the generated text. An effective way to reflect the intended importance of each keyword is to adjust the position of keywords within the text. For example, important keywords such as topic words and eye-catching words can be placed at the beginning of the text to attract the reader's attention, while the keywords for supplementary information can be placed in the middle or later in the text. By controlling the specific position of each keyword according to its importance, it is possible to generate appropriate text for each situation. That is, controlling the specific position of keywords in the generated text is a challenge in terms of reflecting more-specific user intentions and generating texts that attract readers. However, as far as we know, no previous work has tackled this challenge.

	In this paper, we tackle a novel task of controlling keywords and the position of each keyword in text generation. Inspired by previous work that controlled text attributes by using special tokens~\cite{iwama-kano-2019-multiple, lakew-etal-2019-controlling, martin-etal-2020-controllable}, we propose a task-independent method that uses special tokens to control text generation. Specifically, the position of the keyword is specified by providing the model with a special token that represents the target relative position of the keyword (0-10\%, 10-20\%, etc.) and length of target text (20-24 words, 25-29 words, etc.). We use relative positions (rather than absolute positions) because it is more practical to specify relative positions such as ``at the beginning,'' ``in the middle,'' or ``at the end'' of the target text. Moreover, length of the target text is controlled because text length is considered to be one of the important factors that users want to control when considering where to place keywords. During training of the model, the model is provided with control tokens, including keywords randomly extracted from the target text, the positions of each keyword, and the length of the target text. The model is trained with cross-entropy loss in the same manner as conventional text generation; as a result, the model can learn the correspondence between the input control tokens and the target text.
	
	The proposed ``task-independent text-generation-control method'' (``proposed method'' hereafter) was comprehensively evaluated by applying it to summarization and story-generation tasks. The results of the evaluation show that the proposed method can control keywords and their positions in both tasks (Section~\ref{sec:eval_position}). They also show that the proposed method can generate summary texts that are more similar to the gold summary than the baseline, indicating that text closer to the user's intent can be generated (Section~\ref{sec:eval_summarization}). Case studies show that a model specifying keyword position can reflect the user's fine-grained intention (Section~\ref{sec:eval_case}).

	\section{Method}
	\begin{figure}[t]
		\centering
		\includegraphics[clip, width=7.6cm]{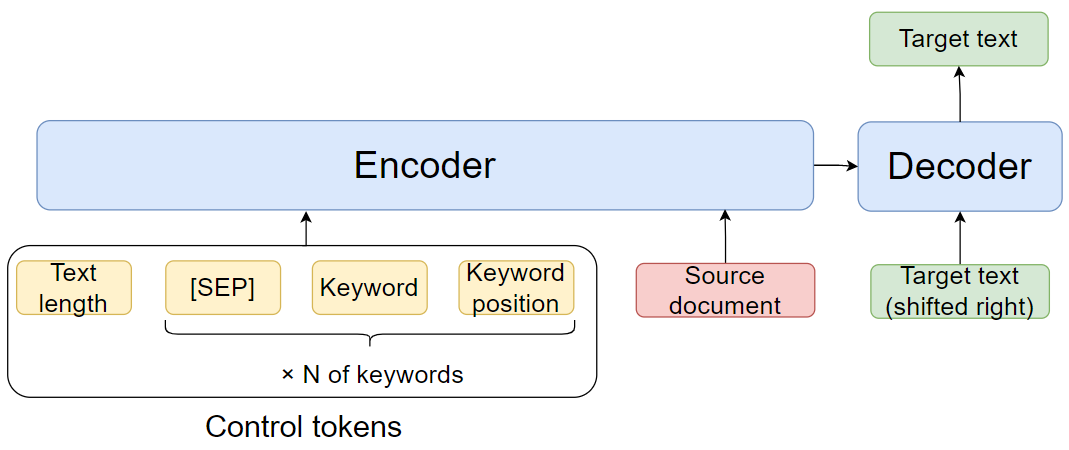}
		\caption{Overview of proposed method. The model is provided with control tokens: keywords in the target text, positions of each keyword, and target-text length to control text generation.}
		\label{fig:model}
	\end{figure}
	
	\subsection{Models}
	A BART model~\cite{lewis-etal-2020-bart} is used for the summarization task, and a GPT model~\cite{radford2019language} is used for the story-generation task. When the BART model is used, the source document is combined with the control tokens (i.e., keywords in the text to be generated, positions of each keyword, and length of the text to be generated) and given to the encoder as shown in Figure~\ref{fig:model}. When the GPT model is used, the control tokens are given to the decoder. As with regular text generation using BART and GPT models, the model is trained to maximize the conditional probabilities $p(y_i|y_{<i},x)$ by using cross-entropy loss, where $y$ denotes the target text and $x$ denotes the input to the model, including the control tokens and the source document used in summarization task.
	
	\subsection{Control tokens}
	\label{method:control_tokens}
	Inspired by existing methods that control text attributes by special tokens~\cite{iwama-kano-2019-multiple, lakew-etal-2019-controlling, martin-etal-2020-controllable}, the model is provided with the position of each keyword and text length as special tokens. For example, if the keyword phrase ``two dogs'' is located in the first 20-30\% of the text and text length is in the range of 50-54 words, ``$\text{\tt [LENGTH50]}\,\text{\tt [SEP]}\,\text{two dogs}\, \text{\tt [POSITION20]}$'' will be given to the model as the control token. Here, $\text{\tt [LENGTH50]}$ and $\text{\tt [POSITION20]}$ are new tokens added to the vocabulary, and the corresponding word embedding is initialized randomly.

	Note that control tokens that represent the oracle information of the target text are given to the model during both training and inference. This setting is appropriate because we aim to generate the intended text by providing additional information to the model. It is also possible that the model automatically determines keywords and their positions (i.e., control tokens are not given to the model), but that approach is left for future work. 
	
	Control tokens are extracted from the target text as follows. More details are given in Appendix~\ref{sec:appendix_control}.
	
	\paragraph{Keywords} Keywords in this paper are not limited to important words in the target text; they can also be any phrase consisting of one to three consecutive words in the target text. For example, from the target text ``Marcia was looking forward to trying hang gliding.'', the phrases ``Marsha'', ``was'', ``looking forward'', ``to trying'', and ``trying hang gliding'' are first extracted as keyword candidates. However, frequent words with little meaning such as ``was'' and ``to trying'' are excluded from the keyword candidates, because they are considered unlikely to be given as keywords by the user. During training, a random number of phrases from the keyword candidates are given to the model as keywords. During inference, the user has the flexibility to give arbitrary keywords to the model. However, for the experiments conducted in this paper, we follow the same approach as during training: the keywords are randomly selected from the keyword candidates and given to the model. 
	
	\paragraph{Keyword Position} The position of each keyword is expressed as a relative position. Specifically, the absolute position of the target keyword when counted from the beginning of the text is divided by the number of words in the text, and the quantized position in units of 10\% are given to the model.

	\paragraph{Text Length} Number of words in the target text (quantized in 5 word units) is given to the model.

	\begin{table}[!t]
		\centering
		\scalebox{0.91}{
			{\tabcolsep=2.0pt
				\begin{tabular}{lcccccc} \hline
					
					\multirow{2}{*}{Control} & \multicolumn{2}{c}{CNN/DM} & \multicolumn{2}{c}{XSum} & \multicolumn{2}{c}{ROCStories} \\
					
					& Include & Pos & Include & Pos & Include & Pos \\ \hline
					\multicolumn{7}{c}{One keyword} \\ \hline
					
					w/o Control & 27.5 & 8.3 &23.4 & 9.4& 0.5& 0.1\\
					Keyword & 71.3 & 18.7& 86.4& 28.7& 53.0& 14.3\\
					+Len & 72.7 & 20.4& 85.8& 30.8& 50.9& 13.5\\
					+Pos & 80.8 & 47.0& \textbf{92.1}& 63.0& 57.2& 27.4\\
					+Pos+Len & \textbf{85.8} & \textbf{48.8} & 91.8& \textbf{64.1} & \textbf{58.8} & \textbf{29.1} \\ \hline
					\multicolumn{7}{c}{Two keywords} \\ \hline
					Keyword & 52.4 & 5.1& 74.1& 14.1& 22.9& 1.6\\
					+Pos+Len & \textbf{75.9} & \textbf{28.6} & \textbf{85.9} & \textbf{46.4} & \textbf{31.1} & \textbf{7.9} \\ \hline
					\multicolumn{7}{c}{Three keywords} \\ \hline
					Keyword & 39.1 & 2.0& 62.5& 9.8& 9.2& 0.3\\
					+Pos+Len & \textbf{70.6} & \textbf{21.8} & \textbf{80.5} & \textbf{37.3}& \textbf{15.5} & \textbf{2.2} \\ \hline
				\end{tabular}
			}
		}
		\caption{Evaluation of the control of keywords and their positions in terms of (i) accuracy of generating text \textbf{Includ}ing all of the target keywords and (ii) accuracy of generating text in which all of the target keywords are placed in each target \textbf{Pos}ition.}
		\label{tab:position_control_result}
	\end{table}
	
	\section{Experiment}
	\begin{table*}[h]
		\centering
		\scalebox{0.88}{
			{\tabcolsep=2.5pt
				\begin{tabular}{lcccccccccc} \hline
					
					Keyword position & \multicolumn{10}{c}{Target-keyword position (relative position)} \\
					in the generated summary & 0-10\% & 10-20\% & 20-30\% & 30-40\% & 40-50\% & 50-60\% & 60-70\% & 70-80\% & 80-90\% & 90-100\%\\ \hline
					\multicolumn{11}{c}{Keyword only Control} \\ \hline
					Correct position & 52.6 & 	23.8 &	14.5 &	9.5& 	9.5 &	9.1 & 	8.7 & 	11.8 &	12.7 &	15.6 
					\\ \hline
					
					\multicolumn{11}{c}{Keyword + Position + Length Control} \\ \hline
					Correct position & \textbf{84.0} & 	\textbf{57.9} &	\textbf{49.1} &	\textbf{41.4} &	\textbf{36.0} &	\textbf{36.2} &	\textbf{33.7} & \textbf{36.0} & \textbf{46.2} &	\textbf{47.9} \\
					Within 10\% diff & 8.1 &	27.5 &	31.9 &	34.4 &	36.1 &	34.1 &	35.5 &	34.3 &	23.3 &	8.9 \\
					Over 10\% diff & 3.2 &	5.3 &	8.3 &	12.8 &	15.1 &	15.5 &	15.1 &	11.4 &	6.7 &	10.9 \\
					Not included & 4.7 & 	9.4 &	10.7 &	11.4 &	12.8 &	14.1 &	15.7 &	18.4 &	23.7 &	32.4 \\ \hline 
				\end{tabular}
			}
		}
		\caption{Detailed evaluation of the control of the keyword and its position in the CNN/DM dataset. For each target relative position of the keyword, the keyword position in the text was classified as (i) located in the target position (\textbf{Correct position}), (ii) located at a positional deviation within 10\% (\textbf{Within 10\% diff}), (iii) located at a positional deviation greater than 10\% (\textbf{Over 10\% diff}), or (iv) not included in the text (\textbf{Not included}).}
		\label{tab:position_control_detail_result}
	\end{table*}
	
	\begin{table}[h]
		\centering
		\scalebox{0.92}{
			{\tabcolsep=4.5pt
				\begin{tabular}{lcccccc} \hline
					\multirow{2}{*}{Control} &
					\multicolumn{3}{c}{CNN/DM} & 
					\multicolumn{3}{c}{XSum} \\
					& R1 & R2 & RL & R1 & R2 & RL \\ \hline
					w/o Control & 43.6 & 20.6 & 40.5 & 44.3 & 21.1 & 36.5 \\
					Keyword & 44.4 & 21.4 & 41.3 & 45.9 & 22.7 & 38.4 \\
					+Len & 45.7 & 22.1 & 42.5 & 47.0 & 23.5 & 39.3 \\ 
					+Pos & 44.9 & 21.9 & 41.8 & 46.7 & 23.6 & 40.2\\
					+Pos+Len & \textbf{46.4} & \textbf{22.8} & \textbf{43.2} & \textbf{47.8} & \textbf{24.5} & \textbf{41.2} \\ \hline 
				\end{tabular}
			}
		}
		\caption{Summarization evaluation by ROUGE score. To reduce the effect on the ROUGE score due to giving target keywords, target keywords are excluded from both the target and generated summaries.} 
		\label{tab:rouge_result}
	\end{table}

	\begin{table*}[!t]
		\centering
		\scalebox{0.680}{
			{\tabcolsep=2.0pt
				\begin{tabular}{c|c} \hline				
					Keyword \& Position & Generated text \\ \hline
					
					true miracle dog (0\%) & \parbox{19.2cm}{``She's a \textbf{true miracle dog} and she deserves a good life,'' foster mother says. Theia was apparently hit by a car and buried in a field. Four days later, she was found emaciated and dirt-covered by a farm worker. A fundraising page has raised more than \$10,000.} \\ \hline
					true miracle dog (40\%) & \parbox{19.2cm}{A stray dog apparently hit by a car and buried in a field survives. The dog, named Theia, is ``a \textbf{true miracle dog},'' her foster owner says. She suffered a dislocated jaw, leg injuries and a caved-in sinus cavity. A fundraising page has raised more than \$10,000.} \\ \hline		
					true miracle dog (90\%) & \parbox{19.2cm}{Stray dog apparently hit by a car and buried in a field survives. Four days later, she is found emaciated and dirt-covered on a farm. She has a dislocated jaw, leg injuries and a caved-in sinus cavity. A good Samaritan has set up a fundraising page for ``the \textbf{true miracle dog}''} \\ \hline
					
					\begin{tabular}{c}hit by (40\%) \\ bully breed mix (60\%)\end{tabular} & \parbox{19.2cm}{A stray dog in Washington State apparently died after being \textbf{hit by} a car. Four days later, she was found emaciated and dirt-covered in a field. Theia, a \textbf{bully breed mix}, has a dislocated jaw, leg injuries and a caved-in sinus cavity.} \\ \hline
					\begin{tabular}{c}hit by (10\%) \\ bully breed mix (0\%)\end{tabular} & \parbox{19.2cm}{Theia, a \textbf{bully breed mix}, was apparently \textbf{hit by} a car and buried in a field. Four days later, she was found emaciated and dirt-covered by a farm worker. She suffered a dislocated jaw, leg injuries and a caved-in sinus cavity. A fundraising page has raised more than \$10,000.} \\ \hline
					
					Keyword \& Position & Generated text (Error cases) \\ \hline
					
					\begin{tabular}{c}buried in (80\%)\end{tabular} & \parbox{19.2cm}{Stray dog apparently hit by a car, apparently killed and \textbf{buried in} a field. Four days later, the dog manages to stagger to a nearby farm. She was found by a farm worker, who took her to a vet for help. Theia is now receiving treatment.} \\ \hline
					
					\begin{tabular}{c}whacked (70\%)\end{tabular} & \parbox{19.2cm}{A stray dog in Washington State apparently died after being hit by a car. Four days later, she was found emaciated and dirt-covered by a farm worker. She suffered a dislocated jaw, leg injuries and a caved-in sinus cavity. A fundraising page has raised more than \$10,000.} \\ \hline
					
				\end{tabular}
			}
		}
		\caption{Examples of generated texts from the CNN/DM dataset}
		\label{tab:keyword_position_example1}
	\end{table*}
	
	\begin{table*}[!t]
		\centering
		\scalebox{0.680}{
			{\tabcolsep=2.0pt
				\begin{tabular}{c|c} \hline
					
					Keyword \& Position  & Generated text \\ \hline
					
					dealership (0\%) & \parbox{19.5cm}{I went to the \textbf{dealership} to look at a car. I was very nervous about buying a car. I asked the salesman if I could try out a certain car. The salesman told me that I could try out a blue car. I drove the blue car home and loved it so much, I bought it.} \\ \hline
					
					dealership (30\%) & \parbox{19.5cm}{I went to the \textbf{dealership} to buy a new car. I was very nervous about the car and the price. I went to the \textbf{dealership} and looked at the price. I decided to buy the car and I was very happy with it. I am glad I went to the \textbf{dealership} because it was a great deal.} \\ \hline
					
					dealership (80\%) & \parbox{19.5cm}{I went to the \textbf{dealership} to look at a car. I was very nervous because I didn't know what I wanted. I asked the salesman if I could pick out a car. The salesman told me that I could get a new car if I paid \$40,000. I drove to the \textbf{dealership} and bought a new car.} \\ \hline

					\begin{tabular}{c}drive to work (50\%) \\ saved enough (70\%)\end{tabular} & \parbox{19.5cm}{I was driving to work one day when I saw a car in the road. I pulled over and asked if I could \textbf{drive to work}. The driver told me that he had \textbf{saved enough} money to buy a new car. I drove to work and paid him back. I drove to work and paid him back and he was very happy.} \\ \hline
					
					\begin{tabular}{c}drive to work (20\%) \\ saved enough (0\%)\end{tabular} & \parbox{19.5cm}{I \textbf{saved enough} money to buy a new car. I went to the car dealership to test drive my new car. I drove the car for a few hours before I left. When I got home, I realized I had forgotten my wallet. I had to \textbf{drive to work} to get my wallet back, but I was happy.} \\ \hline
					
					Keyword \& Position  & Generated text (Error cases) \\ \hline
					
					\begin{tabular}{c}began saving \\money (90\%) \end{tabular} & \parbox{19.5cm}{I was in a hurry to get to work. I had to hurry because I didn't have my car keys. I looked everywhere for my keys, but couldn't find them. I finally found them under my coat, and I was relieved. I was able to grab my keys and walk to work without losing my car keys.} \\ \hline
					
					\begin{tabular}{c}local (0\%) \\ Bob (40\%) \\ enough (60\%) \end{tabular} & \parbox{19.5cm}{\textbf{Bob} was a \textbf{local} handyman. He was hired to fix up a leaky roof on his home. \textbf{Bob} was very handy and did a good job at it. Unfortunately, the roof was too deep and the water would not come out. \textbf{Bob} had to call a \textbf{local} handyman to fix the leaky roof.} \\ \hline
					
				\end{tabular}
			}
		}
		\caption{Examples of generated texts from the ROCStories dataset}
		\label{tab:keyword_position_example2}
	\end{table*}

	\subsection{Experiment setting}
	The proposed method was comprehensively evaluated by applying it to well-established summarization and story-generation tasks. These two tasks have different characteristics. As for summarization, the model extracts information from a source document and compresses it into a short text by using the given control tokens. As for story generation, the model generates text solely on the basis of the given control tokens. For summarization, we used the CNN/DailyMail~\cite{NIPS2015_afdec700} and the XSum~\cite{narayan-etal-2018-dont} dataset and the BART$_{\textsc{LARGE}}$ model (400M parameters)~\cite{lewis-etal-2020-bart}. For story generation, we used the ROCStories~\cite{mostafazadeh-etal-2016-corpus} dataset and the GPT2 model (120M parameters)~\cite{radford2019language}. 
	
	We extract candidate keywords from a target text by using the method described in Section~\ref{method:control_tokens}. During training, no more than three keywords were randomly selected from the keyword candidates for each epoch and given to the model. During inference, one to three keywords randomly selected were given to the model in the experiment of Table~\ref{tab:position_control_result}, and one keyword randomly selected was given to the model in the experiment of Table~\ref{tab:position_control_detail_result} and Table~\ref{tab:rouge_result}. 
	
	In all experiments, training and inference were performed three times, and the mean score was reported. See Appendix~\ref{sec:appendix_exp_detail} for more details on the experimental setup.
	
	\subsection{Evaluation of keyword-position control}
	\label{sec:eval_position}
	Whether the given keywords are placed at given positions was evaluated first in terms of (i) the accuracy of generating text including all target keywords and (ii) the accuracy of generating text in which all target keywords are placed in each target position. As shown in Table~\ref{tab:position_control_result}, the proposed method using special tokens (+Pos and +Pos+Len) can generate text that includes the target keyword at the target position. Providing text-length information along with position information (+Pos+Len) improves the accuracy of keyword-position control, particularly in datasets with long text lengths (CNN/DM and ROCStories). In other words, combining relative position and length information enables the model to place the keywords in appropriate positions. The accuracy of the keyword inclusion is also improved when the keyword position is given. We suspect that the model was informed in advance of where the keywords should be placed; as a result, preventing the model from forgetting to place keywords in the text. It is clear that control accuracy is much lower in the case of story generation compared to summarization. This finding may be because the model is not given the source document and generates text from condition tokens only, so the model is more likely to generate the inappropriate context for keyword inclusion.
	
	A more detailed evaluation is given in Table~\ref{tab:position_control_detail_result}. For each target relative position of the keyword, the keyword position in the text was classified as (i) located in the target position, (ii) located at a positional deviation within 10\%, (iii) located at a positional deviation greater than 10\%, or (iv) not included in the text. It is clear from the results in the table that at all target positions, the accuracy of the keyword-position control is improved compared with that achieved using keyword-only control, and that finding suggests the effectiveness of the proposed method. The results also show a high success rate of keyword inclusion and positional control near the beginning of the text, and a low success rate in the middle and at the end of the text.  This may be because the closer to the end of the text, the more difficult it becomes for the model to generate text that contains the specified keywords while maintaining consistency with the context provided by the preceding words.

	\subsection{Evaluation of summary-content control}
	\label{sec:eval_summarization}
	We show that controlling the text makes it easier for the user to generate the intended text in summarization. The results of the evaluation of summary-content control in summarization by ROUGE score~\cite{lin-2004-rouge} are shown in Table~\ref{tab:rouge_result}. Note that to reduce the effect on the ROUGE score due to giving target keywords, target keywords are excluded from both the target and generated summaries. It is clear from the results in the table that the score is improved by controlling keyword positions and text length, and that finding indicates that such control makes it easier to generate text that is close to the user's intended content.
	
	\subsection{Case study}
	\label{sec:eval_case}
	To better understand how the proposed model behaved, representative examples of generated texts are shown in Table~\ref{tab:keyword_position_example1} and Table~\ref{tab:keyword_position_example2}~\footnote{A source document of summarization, gold texts, and additional examples of generated texts are given in Appendix~\ref{sec:samples_append}.}. In these examples, the keywords and their positions were controlled, although in some examples, the position of the keyword deviates slightly from the target position. It is clear from the table that by assigning different positions for the keywords, it was possible to generate several valid texts with different characteristics. For example, in the example in Table~\ref{tab:keyword_position_example1}, placing the keyword ``true miracle dog'' at the 0\% position generates a text that draws the reader's attention with an eye-catching keyword at the beginning of the text. In contrast, placing that keyword at the 90\% position generates a narrative-style text that describes events in chronological order. It is also clear that even when multiple keywords are given, the order of the keywords can be adjusted by controlling the position of each keyword.

	We also show some cases in which the proposed model produced errors. When a keyword position near the end of the text is specified, the instruction is often ignored, and the keyword is placed in a completely different position or not included in the text. As can be seen from the results in Table~\ref{tab:position_control_detail_result}, the model tends to be poor at placing keywords at the back of the text. 
	
	When comparing the generated text of the summarization task with that of the story generation task, we observed that each of the specified keywords is usually used only once in the generated text of the summarization task, while each of the specified keywords is sometimes used multiple times in the generated text of the story generation task. This may be because the story generation task requires the model to generate text content conditionally only on the specified keywords, causing the model to become overly dependent on them.

	\section{Conclusion}
	A method for controlling keywords and the position of each keyword in generated text is proposed and evaluated experimentally by applying it to two tasks: summarization and story generation. The results of the evaluation show that the proposed method, which uses special tokens, can control the keyword positions in both tasks. They also show that the method can generate summary texts that are more similar to the gold summary than the baseline, and that finding indicates that text closer to the user's intent can be generated.

	\paragraph*{Supplementary Materials Availability Statement} 
	
	\paragraph{Source code}
	\begin{itemize}[noitemsep,topsep=0.8pt]
		\setlength{\parskip}{1.0pt} 
		\setlength{\itemsep}{1.0pt}
		
		\item The source code is available at Github\footnote{\url{https://github.com/ckdjrkffz/controlling_keyword_position}}.
	\end{itemize}
	
	\paragraph{Dataset}
	\begin{itemize}[noitemsep,topsep=0.8pt]
		\setlength{\parskip}{1.0pt}
		\setlength{\itemsep}{1.0pt}
		
		\item The CNN/DM dataset is available at Github\footnote{\url{https://github.com/icml-2020-nlp/semsim/tree/master/datasets}}.
		\item The XSum dataset is available at HuggingFace\footnote{\url{https://huggingface.co/datasets/xsum}}.
		\item The ROCStories dataset is available at here\footnote{\url{https://cs.rochester.edu/nlp/rocstories/}}.
		
	\end{itemize}
	
	\section*{Acknowledgments}
	We thank Atsuki Yamaguchi, Gaku Morio, and Yuta Koreeda for their support.
	This paper was accepted by INLG2023 and is identical to the final version that we submitted~\cite{sasazawa-etal-2023-controlling}. We thank INLG/ACL for allowing us to submit it to the arXiv.

	\bibliography{reference}

\begin{thebibliography}{27}
\expandafter\ifx\csname natexlab\endcsname\relax\def\natexlab#1{#1}\fi

\bibitem[{Chan et~al.(2021)Chan, Chung, and
  Fan}]{DBLP:journals/corr/abs-2112-01012}
Ying{-}Hong Chan, Ho{-}Lam Chung, and Yao{-}Chung Fan. 2021.
\newblock \href {http://arxiv.org/abs/2112.01012} {Improving controllability of
  educational question generation by keyword provision}.
\newblock \emph{CoRR}, abs/2112.01012.

\bibitem[{Chan et~al.(2019)Chan, Chen, Wang, Li, Zhang, Gai, Zhao, and
  Yan}]{chan-etal-2019-stick}
Zhangming Chan, Xiuying Chen, Yongliang Wang, Juntao Li, Zhiqiang Zhang, Kun
  Gai, Dongyan Zhao, and Rui Yan. 2019.
\newblock \href {https://doi.org/10.18653/v1/D19-1501} {Stick to the facts:
  Learning towards a fidelity-oriented {E}-commerce product description
  generation}.
\newblock In \emph{Proceedings of the 2019 Conference on Empirical Methods in
  Natural Language Processing and the 9th International Joint Conference on
  Natural Language Processing (EMNLP-IJCNLP)}, pages 4959--4968, Hong Kong,
  China. Association for Computational Linguistics.

\bibitem[{Dou et~al.(2021)Dou, Liu, Hayashi, Jiang, and
  Neubig}]{dou-etal-2021-gsum}
Zi-Yi Dou, Pengfei Liu, Hiroaki Hayashi, Zhengbao Jiang, and Graham Neubig.
  2021.
\newblock \href {https://doi.org/10.18653/v1/2021.naacl-main.384} {{GS}um: A
  general framework for guided neural abstractive summarization}.
\newblock In \emph{Proceedings of the 2021 Conference of the North American
  Chapter of the Association for Computational Linguistics: Human Language
  Technologies}, pages 4830--4842, Online. Association for Computational
  Linguistics.

\bibitem[{Fan et~al.(2018)Fan, Grangier, and Auli}]{fan-etal-2018-controllable}
Angela Fan, David Grangier, and Michael Auli. 2018.
\newblock \href {https://doi.org/10.18653/v1/W18-2706} {Controllable
  abstractive summarization}.
\newblock In \emph{Proceedings of the 2nd Workshop on Neural Machine
  Translation and Generation}, pages 45--54, Melbourne, Australia. Association
  for Computational Linguistics.

\bibitem[{Fan et~al.(2019)Fan, Lewis, and Dauphin}]{fan-etal-2019-strategies}
Angela Fan, Mike Lewis, and Yann Dauphin. 2019.
\newblock \href {https://doi.org/10.18653/v1/P19-1254} {Strategies for
  structuring story generation}.
\newblock In \emph{Proceedings of the 57th Annual Meeting of the Association
  for Computational Linguistics}, pages 2650--2660, Florence, Italy.
  Association for Computational Linguistics.

\bibitem[{He et~al.(2022)He, Kryscinski, McCann, Rajani, and
  Xiong}]{he-etal-2022-ctrlsum}
Junxian He, Wojciech Kryscinski, Bryan McCann, Nazneen Rajani, and Caiming
  Xiong. 2022.
\newblock \href {https://aclanthology.org/2022.emnlp-main.396} {{CTRL}sum:
  Towards generic controllable text summarization}.
\newblock In \emph{Proceedings of the 2022 Conference on Empirical Methods in
  Natural Language Processing}, pages 5879--5915, Abu Dhabi, United Arab
  Emirates. Association for Computational Linguistics.

\bibitem[{Hermann et~al.(2015)Hermann, Kocisky, Grefenstette, Espeholt, Kay,
  Suleyman, and Blunsom}]{NIPS2015_afdec700}
Karl~Moritz Hermann, Tomas Kocisky, Edward Grefenstette, Lasse Espeholt, Will
  Kay, Mustafa Suleyman, and Phil Blunsom. 2015.
\newblock \href
  {https://proceedings.neurips.cc/paper/2015/file/afdec7005cc9f14302cd0474fd0f3c96-Paper.pdf}
  {Teaching machines to read and comprehend}.
\newblock In \emph{Advances in Neural Information Processing Systems},
  volume~28. Curran Associates, Inc.

\bibitem[{Iwama and Kano(2019)}]{iwama-kano-2019-multiple}
Kango Iwama and Yoshinobu Kano. 2019.
\newblock \href {https://doi.org/10.18653/v1/W19-8612} {Multiple news headlines
  generation using page metadata}.
\newblock In \emph{Proceedings of the 12th International Conference on Natural
  Language Generation}, pages 101--105, Tokyo, Japan. Association for
  Computational Linguistics.

\bibitem[{Jain et~al.(2017)Jain, Agrawal, Mishra, Sukhwani, Laha, and
  Sankaranarayanan}]{DBLP:journals/corr/JainAMSLS17}
Parag Jain, Priyanka Agrawal, Abhijit Mishra, Mohak Sukhwani, Anirban Laha, and
  Karthik Sankaranarayanan. 2017.
\newblock \href {http://arxiv.org/abs/1707.05501} {Story generation from
  sequence of independent short descriptions}.
\newblock \emph{CoRR}, abs/1707.05501.

\bibitem[{Kingma and Ba(2015)}]{DBLP:journals/corr/KingmaB14}
Diederik~P. Kingma and Jimmy Ba. 2015.
\newblock \href {http://arxiv.org/abs/1412.6980} {Adam: {A} method for
  stochastic optimization}.
\newblock In \emph{3rd International Conference on Learning Representations,
  {ICLR} 2015, San Diego, CA, USA, May 7-9, 2015, Conference Track
  Proceedings}.

\bibitem[{Lakew et~al.(2019)Lakew, Di~Gangi, and
  Federico}]{lakew-etal-2019-controlling}
Surafel~Melaku Lakew, Mattia Di~Gangi, and Marcello Federico. 2019.
\newblock \href {https://aclanthology.org/2019.iwslt-1.31} {Controlling the
  output length of neural machine translation}.
\newblock In \emph{Proceedings of the 16th International Conference on Spoken
  Language Translation}, Hong Kong. Association for Computational Linguistics.

\bibitem[{Lee et~al.(2018)Lee, Mansimov, and Cho}]{lee-etal-2018-deterministic}
Jason Lee, Elman Mansimov, and Kyunghyun Cho. 2018.
\newblock \href {https://doi.org/10.18653/v1/D18-1149} {Deterministic
  non-autoregressive neural sequence modeling by iterative refinement}.
\newblock In \emph{Proceedings of the 2018 Conference on Empirical Methods in
  Natural Language Processing}, pages 1173--1182, Brussels, Belgium.
  Association for Computational Linguistics.

\bibitem[{Lewis et~al.(2020)Lewis, Liu, Goyal, Ghazvininejad, Mohamed, Levy,
  Stoyanov, and Zettlemoyer}]{lewis-etal-2020-bart}
Mike Lewis, Yinhan Liu, Naman Goyal, Marjan Ghazvininejad, Abdelrahman Mohamed,
  Omer Levy, Veselin Stoyanov, and Luke Zettlemoyer. 2020.
\newblock \href {https://doi.org/10.18653/v1/2020.acl-main.703} {{BART}:
  Denoising sequence-to-sequence pre-training for natural language generation,
  translation, and comprehension}.
\newblock In \emph{Proceedings of the 58th Annual Meeting of the Association
  for Computational Linguistics}, pages 7871--7880, Online. Association for
  Computational Linguistics.

\bibitem[{Lin(2004)}]{lin-2004-rouge}
Chin-Yew Lin. 2004.
\newblock \href {https://aclanthology.org/W04-1013} {{ROUGE}: A package for
  automatic evaluation of summaries}.
\newblock In \emph{Text Summarization Branches Out}, pages 74--81, Barcelona,
  Spain. Association for Computational Linguistics.

\bibitem[{Martin et~al.(2020)Martin, de~la Clergerie, Sagot, and
  Bordes}]{martin-etal-2020-controllable}
Louis Martin, {\'E}ric de~la Clergerie, Beno{\^\i}t Sagot, and Antoine Bordes.
  2020.
\newblock \href {https://aclanthology.org/2020.lrec-1.577} {Controllable
  sentence simplification}.
\newblock In \emph{Proceedings of the 12th Language Resources and Evaluation
  Conference}, pages 4689--4698, Marseille, France. European Language Resources
  Association.

\bibitem[{Mostafazadeh et~al.(2016)Mostafazadeh, Chambers, He, Parikh, Batra,
  Vanderwende, Kohli, and Allen}]{mostafazadeh-etal-2016-corpus}
Nasrin Mostafazadeh, Nathanael Chambers, Xiaodong He, Devi Parikh, Dhruv Batra,
  Lucy Vanderwende, Pushmeet Kohli, and James Allen. 2016.
\newblock \href {https://doi.org/10.18653/v1/N16-1098} {A corpus and cloze
  evaluation for deeper understanding of commonsense stories}.
\newblock In \emph{Proceedings of the 2016 Conference of the North {A}merican
  Chapter of the Association for Computational Linguistics: Human Language
  Technologies}, pages 839--849, San Diego, California. Association for
  Computational Linguistics.

\bibitem[{Narayan et~al.(2018)Narayan, Cohen, and
  Lapata}]{narayan-etal-2018-dont}
Shashi Narayan, Shay~B. Cohen, and Mirella Lapata. 2018.
\newblock \href {https://doi.org/10.18653/v1/D18-1206} {Don{'}t give me the
  details, just the summary! topic-aware convolutional neural networks for
  extreme summarization}.
\newblock In \emph{Proceedings of the 2018 Conference on Empirical Methods in
  Natural Language Processing}, pages 1797--1807, Brussels, Belgium.
  Association for Computational Linguistics.

\bibitem[{Ni and McAuley(2018)}]{ni-mcauley-2018-personalized}
Jianmo Ni and Julian McAuley. 2018.
\newblock \href {https://doi.org/10.18653/v1/P18-2112} {Personalized review
  generation by expanding phrases and attending on aspect-aware
  representations}.
\newblock In \emph{Proceedings of the 56th Annual Meeting of the Association
  for Computational Linguistics (Volume 2: Short Papers)}, pages 706--711,
  Melbourne, Australia. Association for Computational Linguistics.

\bibitem[{Radford et~al.(2018)Radford, Wu, Child, Luan, Amodei, and
  Sutskever}]{radford2019language}
Alec Radford, Jeffrey Wu, Rewon Child, David Luan, Dario Amodei, and Ilya
  Sutskever. 2018.
\newblock \href
  {https://d4mucfpksywv.cloudfront.net/better-language-models/language-models.pdf}
  {Language models are unsupervised multitask learners}.

\bibitem[{Sasazawa et~al.(2023)Sasazawa, Morishita, Ozaki, Imaichi, and
  Sogawa}]{sasazawa-etal-2023-controlling}
Yuichi Sasazawa, Terufumi Morishita, Hiroaki Ozaki, Osamu Imaichi, and Yasuhiro
  Sogawa. 2023.
\newblock \href {https://aclanthology.org/2023.inlg-main.29} {Controlling
  keywords and their positions in text generation}.
\newblock In \emph{Proceedings of the 16th International Natural Language
  Generation Conference}, pages 407--413, Prague, Czechia. Association for
  Computational Linguistics.

\bibitem[{Shao et~al.(2021)Shao, Wang, Lin, Zhang, Zhang, Ji, and
  Abdelzaher}]{10.1145/3442381.3449838}
Huajie Shao, Jun Wang, Haohong Lin, Xuezhou Zhang, Aston Zhang, Heng Ji, and
  Tarek Abdelzaher. 2021.
\newblock \href {https://doi.org/10.1145/3442381.3449838} {Controllable and
  diverse text generation in e-commerce}.
\newblock In \emph{Proceedings of the Web Conference 2021}, WWW '21, page
  2392^^e2^^80^^932401, New York, NY, USA. Association for Computing Machinery.

\bibitem[{Su et~al.(2021)Su, Vandyke, Wang, Fang, and
  Collier}]{su-etal-2021-plan-generate}
Yixuan Su, David Vandyke, Sihui Wang, Yimai Fang, and Nigel Collier. 2021.
\newblock \href {https://doi.org/10.18653/v1/2021.findings-emnlp.76}
  {Plan-then-generate: Controlled data-to-text generation via planning}.
\newblock In \emph{Findings of the Association for Computational Linguistics:
  EMNLP 2021}, pages 895--909, Punta Cana, Dominican Republic. Association for
  Computational Linguistics.

\bibitem[{Szegedy et~al.(2016)Szegedy, Vanhoucke, Ioffe, Shlens, and
  Wojna}]{Szegedy-2016-Rethinking}
Christian Szegedy, Vincent Vanhoucke, Sergey Ioffe, Jon Shlens, and Zbigniew
  Wojna. 2016.
\newblock \href {https://doi.org/10.1109/CVPR.2016.308} {Rethinking the
  inception architecture for computer vision}.
\newblock In \emph{2016 IEEE Conference on Computer Vision and Pattern
  Recognition (CVPR)}, pages 2818--2826.

\bibitem[{Yao et~al.(2019)Yao, Peng, Weischedel, Knight, Zhao, and
  Yan}]{10.1609/aaai.v33i01.33017378}
Lili Yao, Nanyun Peng, Ralph Weischedel, Kevin Knight, Dongyan Zhao, and Rui
  Yan. 2019.
\newblock \href {https://doi.org/10.1609/aaai.v33i01.33017378} {Plan-and-write:
  Towards better automatic storytelling}.
\newblock In \emph{Proceedings of the Thirty-Third AAAI Conference on
  Artificial Intelligence and Thirty-First Innovative Applications of
  Artificial Intelligence Conference and Ninth AAAI Symposium on Educational
  Advances in Artificial Intelligence}, AAAI'19/IAAI'19/EAAI'19. AAAI Press.

\bibitem[{Yoon et~al.(2020)Yoon, Yeo, Jeong, Yi, and
  Kang}]{DBLP:journals/corr/abs-2002-07767}
Wonjin Yoon, Yoon~Sun Yeo, Minbyul Jeong, Bong{-}Jun Yi, and Jaewoo Kang. 2020.
\newblock \href {http://arxiv.org/abs/2002.07767} {Learning by semantic
  similarity makes abstractive summarization better}.
\newblock \emph{CoRR}, abs/2002.07767.

\bibitem[{Zhang and Zhu(2021)}]{10.1145/3442381.3449876}
Zhiling Zhang and Kenny Zhu. 2021.
\newblock \href {https://doi.org/10.1145/3442381.3449876} {Diverse and specific
  clarification question generation with keywords}.
\newblock In \emph{Proceedings of the Web Conference 2021}, WWW '21, page
  3501^^e2^^80^^933511, New York, NY, USA. Association for Computing Machinery.

\bibitem[{Zhu et~al.(2018)Zhu, Lu, Zheng, Guo, Zhang, Wang, and
  Yu}]{10.1145/3209978.3210080}
Yaoming Zhu, Sidi Lu, Lei Zheng, Jiaxian Guo, Weinan Zhang, Jun Wang, and Yong
  Yu. 2018.
\newblock \href {https://doi.org/10.1145/3209978.3210080} {Texygen: A
  benchmarking platform for text generation models}.
\newblock In \emph{The 41st International ACM SIGIR Conference on Research \&
  Development in Information Retrieval}, SIGIR '18, page 1097^^e2^^80^^931100,
  New York, NY, USA. Association for Computing Machinery.

\end{thebibliography}
	\bibliographystyle{acl_natbib}
	
	\clearpage
	
	\appendix
	
	\begin{table*}[!t]
		\centering
		{\tabcolsep=3.5pt
			\begin{tabular}{lccccc} \hline
				Datasets & \#train & \#dev & \#test & \#source document tokens & \#target text tokens \\ \hline
				CNN/DM & 287,227 & 13,368 & 11,490 & 777.6 (399.5) & 57.9 (24.6) \\
				XSum & 204,045 & 11,332 & 11,334 & 433.1 (355.8) & 23.2 (5.8) \\ 
				ROCStories & 78,528 & 9,816 & 9,817 & -- & 49.8 (9.4) \\ \hline
			\end{tabular}
		}
		\caption{Dataset statistics: number of training data, number of development data, number of test data, number of words in source document and its standard deviation, and number of words in target text and its standard deviation.}
		\label{tab:data_statistics}
	\end{table*}

	\section{Experiment settings}
	~\label{sec:appendix_exp_detail}
	\subsection{Hyper-parameter}
	~\label{sec:appendix_param}
	Adam~\cite{DBLP:journals/corr/KingmaB14}, with $\beta_1 = 0.9$, $\beta_2 = 0.999$, $\epsilon=10^{-6}$, and L2 regularization factor $0.01$, was used as the optimizer. We use linear warmup over the first 6\% of the training steps and linear decayed. The dropout rate was 0.1, a batch size was 32, and label smoothing~\cite{Szegedy-2016-Rethinking} was 0.1.
	
	\paragraph{Summarization task} The learning rate was set to $2 \times 10^{-5}$ for both models. However, since the weights of the newly added special tokens were trained from the initialization state, the learning rate for the word embedding weights was set higher, namely, $1 \times 10^{-3}$. Number of epochs was set to 10. During inference, the summary text was generated by using a beam search. For the CNN/DM dataset, the number of beams was 4, and the length penalty was 2.0. For the XSum dataset, the number of beams was 6, and the length penalty was 1.0. Maximum number of tokens for the source document was set to 1024, and maximum number of tokens for the summary text was set to 128. Any text with a higher number of tokens was truncated at the end.
	
	\paragraph{Story generation task} The learning rate was set to $2 \times 10^{-5}$, and the learning rate for word embedding weights was set to $1 \times 10^{-3}$. Number of epochs is 30. During generation, texts were generated by top-p sampling (where $p=0.95$). The temperature was set to 0.1. Maximum number of tokens for the target text was set to 128, and any text with a higher number of tokens was truncated at the end.
	
	\paragraph{Hyper-parameter search} The batch size was selected from $\{32, 64, 128, 256\}$. The learning rate was selected from $\{1 \times 10^{-5}, 2 \times 10^{-5}, 5 \times 10^{-5}, 1 \times 10^{-4}, and 2 \times 10^{-4}\}$. Number of epochs was selected from $\{3, 5, 10, 15, 20\}$ for summarization and $\{10, 20, 30, 60, 100\}$ for story generation. Due to the large search area, we used the manual search to determine parameters. To determine the hyper-parameters, we used the accuracy of the keyword-position control listed in Table~\ref{tab:position_control_result}.
	
	\subsection{Environment}
	\label{sec:appendix_environment}
	We used $4 \times$ V100 GPUs for training and $1 \times$ V100 GPU for inference. Training on the CNN/DM, XSum, and ROCStories datasets took 30, 40, and 4 hours, respectively, and inference took less than 1 hour for all datasets. 
	
	\subsection{Dataset and Control tokens}
	\label{sec:appendix_control}
	The statistics of the dataset are shown in Table~\ref{tab:data_statistics}. For the CNN/DM dataset, we followed the data split proposed by ~\citet{DBLP:journals/corr/abs-2002-07767}. For the XSum dataset, we followed the data split proposed by ~\citet{narayan-etal-2018-dont}. For ROCStories, we split the data into 8:1:1 for training, development, and test sets. Control tokens (keywords, each keyword position, and text length) were extracted from the target text and given to the model for training. Word tokenization of the text was done by NLTK library\footnote{\url{https://github.com/nltk/nltk/blob/develop/nltk/tokenize/__init__.py}} to obtain keywords and text length. Note that the model receives the text tokenized into subwords. Therefore, the number of tokens the model receives differs from the pre-calculated text length.
	
	In training, the model was provided with all control tokens (keywords, keyword position, and text length) each with a certain probability. The trained model was then used to generate texts under four different settings (Keyword, +Len, +Pos, and +Len+Pos in Table~\ref{tab:position_control_result}). In this way, a single model can perform inference experiments in multiple settings in a manner that lowers the cost of the experiments. Preliminary experiments showed that performance of each model is almost the same as when each model is trained for each inference setting. However, for regular text generation that does not use control tokens (``w/o Control'' in Table~\ref{tab:position_control_result}), model training and inference were performed separately from the model described above.
	
	\paragraph{Keywords} A sequence consisting of one to three consecutive words was obtained from the target text as keyword candidates. Phrases whose first word is a stop word or a frequent word were excluded from the keyword candidates because they are considered unlikely to be given as keywords by the user. During training, no more than three keywords were randomly selected from the keyword candidates for each epoch and given to the model. During inference, one to three keywords were given to the model in the experiment of Table ~\ref{tab:position_control_result}, and one keyword was given to the model in the experiment of Table ~\ref{tab:position_control_detail_result} and Table~\ref{tab:rouge_result}. In both training and inference, when multiple keywords are selected from candidate keywords, they are selected so that one keyword is not part of another (e.g., ``looking'' vs ``looking forward'').
	
	\paragraph{Keyword Position} Relative positions of the above keywords in the target text were obtained and given to the model. However, for each keyword with a probability of 10\%, the keyword position was not given to the model, and only the keyword was given during training.
	
	\paragraph{Text Length} Word length of the target text was obtained and given to the model as text length. Also, with a probability of 10\%, text length was not given to the model during training.
	
	\section{Generating diverse texts}
	\begin{table}[!t]
		\centering
		\scalebox{1.0}{
			{\tabcolsep=1.5pt
				\begin{tabular}{lccc} \hline
					\multirow{2}{*}{Control} &
					CNN/DM & XSum & ROCStories \\
					& S-BLEU↓ & S-BLEU↓ & S-BLEU↓\\ \hline
					w/o Control & 96.2 & 92.2 & -- \\
					Keyword & 96.6 & 92.8 & 71.2 \\
					+Pos+Len & \textbf{85.9} & \textbf{83.1} & \textbf{56.9} \\ \hline 
				\end{tabular}
			}
		}
		\caption{Evaluation of diversity of generated texts using the Self-BLEU metric}
		\label{tab:diversity_result}
	\end{table}
	
	Here, we show that by controlling keyword positions, it is possible to generate a variety of texts from a specific keyword. The ability to generate diverse texts will enable users to select their intended text from multiple generated texts. When normal generation (``w/o Control'') and with keyword generation, 10 different texts are generated from a single input\footnote{As indicated in Appendix~\ref{sec:appendix_param}, beam search was used for summarization, and top-p sampling was used for story generation.}. When keyword position is used, the model is given 10 different keyword positions for one particular keyword, and 10 different texts are generated. For each generated text, the diversity is evaluated with the Self-BLEU~\cite{10.1145/3209978.3210080} score. The results in Table~\ref{tab:diversity_result} show that the generated text diversity is improved by providing a variety of keyword positions. In particular, beam search was used to generate multiple texts in summarization, and it resulted in very low diversity in the generated texts. This reduction in diversity, however, was mitigated by controlling the keyword position. The examples of generated texts in Table~\ref{tab:keyword_position_example1} show that controlling keyword position produces a variety of valid texts from a particular keyword.
	
	\section{Generation specifying random positions}
	\begin{table}[!t]
		\centering
		\scalebox{0.85}{
			{\tabcolsep=3.0pt
				\begin{tabular}{lcccccc} \hline
					
					\multirow{2}{*}{Control} & \multicolumn{2}{c}{CNN/DM} & \multicolumn{2}{c}{XSum} & \multicolumn{2}{c}{ROCStories} \\
					& Include & Pos & Include & Pos & Include & Pos \\ \hline
					\multicolumn{7}{c}{Specifying oracle Position} \\ \hline
					+Pos+Len & 85.8 & 48.8 & 91.8 & 64.1 & 58.8 & 29.1 \\ \hline
					\multicolumn{7}{c}{Specifying random Position} \\ \hline
					+Pos+Len & 84.2 & 36.6 & 89.0 & 43.1 & 58.7 & 23.8 \\ \hline
				\end{tabular}
			}
		}
		\caption{Comparison between specifying oracle keyword positions and random keyword positions. The keywords given to the model are single oracle keywords extracted from the target text, and they are the same in both settings.}
		\label{tab:random_position_control_result}
	\end{table}
	
	In section~\ref{sec:eval_position}, we showed that it is possible to generate texts by specifying oracle keyword positions extracted from target texts. We also show that it is possible to generate text by specifying arbitrary keyword positions. Specifically, we compare the accuracy of position control when oracle keyword positions are given and that when randomly selected keyword positions are given. Note that the keywords given to the model are single oracle keywords extracted from the target text, and they are same in both settings. 
	
	As shown by the results in Table~\ref{tab:random_position_control_result}, position control is still possible when a random position is specified. However, when random positions are specified, the accuracy of keyword inclusion is slightly lower, and the accuracy of keyword-position control is significantly lower. That result can be explained by the fact that keyword positions that are difficult to satisfied can be specified. For example, a keyword originally used at the end of the text is difficult to place at the beginning of the text.
	
	\section{Generated samples}
	\label{sec:samples_append}
	Examples of texts generated by the proposed method as given in Table~\ref{tab:keyword_position_example_appendix1}, Table~\ref{tab:keyword_position_example_appendix2}, and Table~\ref{tab:keyword_position_example_appendix3}.
	
	\section{Limitations}
	\paragraph{Using oracle information} In this paper, text generation is controlled by providing the model with control tokens extracted from the target text. The improved accuracy of keyword inclusion and position control shown in our experiments is due to this additional information, not to improved performance of the model itself. Because the goal of this paper is to enable users to control the model by providing additional information such as keywords and positions, this experiment design is not a mistake. However, selecting appropriate keywords and placing them in the appropriate positions without relying on oracle information is one of the challenges for future work.
	
	\paragraph{Using length information} The proposed method requires that the model be given a target text length, which may impose an extra burden on the user in practical terms. Experimental results showed that length information itself is not essential for controlling relative position, but it is one key to improving performance. ~\citet{lee-etal-2018-deterministic} proposed a method for predicting target-text length from the source document used in machine translation. By incorporating this method, it may be possible to control the relative positions of keywords without providing additional length information.
	
	\paragraph{Insufficient performance} The experiment results in Table~\ref{tab:position_control_result} show that accuracy of keyword inclusion and keyword-position control is low, especially in the case of story generation. The reason for this low accuracy may be that the model does not generate the appropriate context for the inclusion of keywords because the source document is not given. In the summarization, accuracy of the keyword-position control is also far from 100\% control accuracy. Since it is possible to extract only desirable text from the generated text, the control need not necessarily succeed 100\% of the time. However, if success rate of the control is improved, efficiency of text generation will improve. A deeper investigation of the cause of poor performance and improvement in control accuracy are challenges for future work. One idea to improve performance of story generation is to give the model several words at the beginning of the text, and this approach may make it easier for the model to generate the appropriate context. 
	
	\clearpage

	\begin{table*}[!t]
		\centering
		\scalebox{0.80}{
			{\tabcolsep=2.0pt
				\begin{tabular}{c|c} \hline
					\multicolumn{2}{c}{Source document} \\ \hline
					\multicolumn{2}{c}{\parbox{19.5cm}{Never mind cats having nine lives. A stray pooch in Washington State has used up at least three of her own after being hit by a car, apparently whacked on the head with a hammer in a misguided mercy killing and then buried in a field -- only to survive. That\'s according to Washington State University, where the dog -- a friendly white-and-black bully breed mix now named Theia -- has been receiving care at the Veterinary Teaching Hospital. Four days after her apparent death, the dog managed to stagger to a nearby farm, dirt-covered and emaciated, where she was found by a worker who took her to a vet for help. She was taken in by Moses Lake, Washington, resident Sara Mellado. ``Considering everything that she\'s been through, she\'s incredibly gentle and loving,'' Mellado said, according to WSU News. ``She\'s a true miracle dog and she deserves a good life.'' Theia is only one year old but the dog\'s brush with death did not leave her unscathed. She suffered a dislocated jaw, leg injuries and a caved-in sinus cavity -- and still requires surgery to help her breathe. The veterinary hospital\'s Good Samaritan Fund committee awarded some money to help pay for the dog\'s treatment, but Mellado has set up a fundraising page to help meet the remaining cost of the dog\'s care. She\'s also created a Facebook page to keep supporters updated. Donors have already surpassed the \$10,000 target, inspired by Theia\'s tale of survival against the odds. On the fundraising page, Mellado writes, ``She is in desperate need of extensive medical procedures to fix her nasal damage and reset her jaw. I agreed to foster her until she finally found a loving home.'' She is dedicated to making sure Theia gets the medical attention she needs, Mellado adds, and wants to ``make sure she gets placed in a family where this will never happen to her again!'' Any additional funds raised will be ``paid forward'' to help other animals. Theia is not the only animal to apparently rise from the grave in recent weeks. A cat in Tampa, Florida, found seemingly dead after he was hit by a car in January, showed up alive in a neighbor\'s yard five days after he was buried by his owner. The cat was in bad shape, with maggots covering open wounds on his body and a ruined left eye, but remarkably survived with the help of treatment from the Humane Society.}} \\ \hline
					
					\multicolumn{2}{c}{Gold text} \\ \hline
					\multicolumn{2}{c}{\parbox{19.5cm}{Theia, a bully breed mix, was apparently hit by a car, whacked with a hammer and buried in a field. ``She\'s a true miracle dog and she deserves a good life,'' says Sara Mellado, who is looking for a home for Theia.}} \\ \hline
					
					Keyword \& Position & Generated text \\ \hline
					
					w/o Control & \parbox{15.5cm}{A dog apparently hit by a car and buried in a field appears to have survived. Four days after her apparent death, the dog stagger to a nearby farm. She was found by a worker who took her to a vet for help. A fundraising page has raised more than \$10,000 for the dog's care.} \\ \hline
					
					\begin{tabular}{c}true miracle dog \\ (w/o Position Control)\end{tabular} & \parbox{15.5cm}{Stray dog apparently hit by a car and buried in a field survives. She was found emaciated and dirt-covered by a farm worker four days later. The dog, named Theia, has a dislocated jaw, leg injuries and a caved-in sinus cavity. A fundraising page has raised more than \$10,000.} \\ \hline

					true miracle dog (0\%) & \parbox{15.5cm}{``She's a \textbf{true miracle dog} and she deserves a good life,'' foster mother says. Theia was apparently hit by a car and buried in a field. Four days later, she was found emaciated and dirt-covered by a farm worker. A fundraising page has raised more than \$10,000.} \\ \hline
					true miracle dog (40\%) & \parbox{15.5cm}{A stray dog apparently hit by a car and buried in a field survives. The dog, named Theia, is ``a \textbf{true miracle dog},'' her foster owner says. She suffered a dislocated jaw, leg injuries and a caved-in sinus cavity. A fundraising page has raised more than \$10,000.} \\ \hline		
					true miracle dog (90\%) & \parbox{15.5cm}{Stray dog apparently hit by a car and buried in a field survives. Four days later, she is found emaciated and dirt-covered on a farm. She has a dislocated jaw, leg injuries and a caved-in sinus cavity. A good Samaritan has set up a fundraising page for ``the \textbf{true miracle dog}``} \\ \hline
					\begin{tabular}{c}hit by (40\%) \\ bully breed mix (60\%)\end{tabular} & \parbox{15.5cm}{A stray dog in Washington State apparently died after being \textbf{hit by} a car. Four days later, she was found emaciated and dirt-covered in a field. Theia, a \textbf{bully breed mix}, has a dislocated jaw, leg injuries and a caved-in sinus cavity.} \\ \hline
					
					\begin{tabular}{c}hit by (10\%) \\ bully breed mix (0\%)\end{tabular} & \parbox{15.5cm}{Theia, a \textbf{bully breed mix}, was apparently \textbf{hit by} a car and buried in a field. Four days later, she was found emaciated and dirt-covered by a farm worker. She suffered a dislocated jaw, leg injuries and a caved-in sinus cavity. A fundraising page has raised more than \$10,000.} \\ \hline			
					
					\begin{tabular}{c}apparently (10\%) \\ field (40\%) \\ miracle dog (60\%)\end{tabular} & \parbox{15.5cm}{A stray dog in Washington State \textbf{apparently} died after being hit by a car. Four days later, the dog was found in a \textbf{field}, dirt-covered and emaciated. ``She's a true \textbf{miracle dog} and she deserves a good life,'' the dog's foster owner says.} \\ \hline
					
					Keyword \& Position & Generated text (Error cases) \\ \hline
					
					\begin{tabular}{c}buried in (80\%)\end{tabular} & \parbox{15.5cm}{Stray dog apparently hit by a car, apparently killed and \textbf{buried in} a field. Four days later, the dog manages to stagger to a nearby farm. She was found by a farm worker, who took her to a vet for help. Theia is now receiving treatment.} \\ \hline
					
					\begin{tabular}{c}whacked (70\%)\end{tabular} & \parbox{15.5cm}{A stray dog in Washington State apparently died after being hit by a car. Four days later, she was found emaciated and dirt-covered by a farm worker. She suffered a dislocated jaw, leg injuries and a caved-in sinus cavity. A fundraising page has raised more than \$10,000.} \\ \hline
					
				\end{tabular}
			}
		}
		\caption{Examples of generated texts from the CNN/DM dataset. This table is the complete version of Table~\ref{tab:keyword_position_example1} with source document, gold summary, and additional examples.}
		\label{tab:keyword_position_example_appendix1}
	\end{table*}
	
	\begin{table*}[!t]
		\centering
		\scalebox{0.80}{
			{\tabcolsep=2.0pt
				\begin{tabular}{c|c} \hline
					\multicolumn{2}{c}{Source document} \\ \hline
					\multicolumn{2}{c}{\parbox{19.5cm}{``I\'m really looking forward to it - the home of Scottish football,'' said Rodgers ahead of his maiden visit. ``I hear the pitch is good, a nice big pitch suits the speed in our team and our intensity. ``The technical area goes right out to the end of the pitch, but you might need a taxi to get back to your staff.'' This will be Rodgers\' second taste of the Old Firm derby and his experience of the fixture got off to a great start with a 5-1 league victory at Celtic Park last month. ``It was a brilliant performance by the players in every aspect,'' he recalled. ``Obviously this one is on a neutral ground, but we\'ll be looking to have a similar performance. ``We\'ll be prepared and focused. We know it\'s going to be a tough game. We anticipated that the last time.'' Rodgers is also aware Celtic\'s visit to Hampden last season in the Scottish Cup semi-final against Rangers ended in defeat. ``The last experience there wasn\'t so good for Celtic,'' he said. ``But it\'s a different team with a different mentality and we\'ll look to take that mentality into the game.'' Rodgers lost two semi-finals in his time at Liverpool and is aiming to make it third time lucky at the club he joined in the summer. ``You have to perform - that\'s what\'s important and if you get little bit of luck along the way then hopefully you\'ll get the result that you want.'' said the Northern Irishman. ``So, for us, it\'s really looking at our performance level, which in the main has been at a really high level. ``My focus is always on my own team and, if we impose our style of play, it can give a lot of problems to opponents.'' Unlike predecessor Ronny Deila, Rodgers would not be drawn on this semi-final being a step on the way to a potential domestic treble. ``It\'s very, very difficult to achieve,'' he said. ``There\'s been great managers here in the past that haven\'t achieved that. ``But it\'s the first competition where we\'ve a chance to win a trophy and I\'ve always said that it\'s a priority for us because it\'s the first one. ``Our idea was to qualify for the Champions League, which we did, to make a good start in the league, which we\'ve done, and then look to get to the League Cup final. ``Unless you can do that then you don\'t have the chance to win the trophies.}} \\ \hline
					
					\multicolumn{2}{c}{Gold text} \\ \hline
					\multicolumn{2}{c}{\parbox{19.5cm}{Manager Brendan Rodgers is sure Celtic can exploit the wide open spaces of Hampden when they meet Rangers in Sunday's League Cup semi-final.}} \\ \hline
					
					Keyword \& Position  & Generated text \\ \hline
					
					w/o Control & \parbox{15.5cm}{Celtic manager Brendan Rodgers is relishing his first visit to Hampden Park as he prepares for Saturday's Scottish Cup semi-final against Rangers.} \\ \hline
					
					\begin{tabular}{c}Hampden \\ (w/o Position Control)\end{tabular} & \parbox{15.5cm}{Celtic manager Brendan Rodgers is relishing the prospect of playing his first Scottish Cup semi-final against Rangers at \textbf{Hampden} Park on Saturday.} \\ \hline

					Hampden (0\%) & \parbox{15.5cm}{\textbf{Hampden} Park will be the \``home of Scottish football\'' when Celtic take on Rangers in the Scottish Cup semi-final, says manager Brendan Rodgers.} \\ \hline
					Hampden (40\%) & \parbox{15.5cm}{New Celtic manager Brendan Rodgers is relishing the prospect of playing at \textbf{Hampden} Park for the first time in the Scottish Cup semi-final against Rangers.} \\ \hline		
					Hampden (80\%) & \parbox{15.5cm}{Celtic manager Brendan Rodgers is relishing the prospect of playing his first Scottish Cup semi-final against Rangers at \textbf{Hampden} Park.''} \\ \hline
					\begin{tabular}{c}Brendan (20\%) \\ meet Rangers (50\%)\end{tabular} & \parbox{15.5cm}{Celtic manager \textbf{Brendan} Rodgers is relishing the opportunity to \textbf{meet Rangers} in the Scottish Cup semi-final at Hampden.} \\ \hline
					
					\begin{tabular}{c}Brendan (80\%) \\ meet Rangers (0\%)\end{tabular} & \parbox{15.5cm}{Celtic will \textbf{meet Rangers} in the Scottish Cup semi-final at Hampden Park for the first time on Saturday, but manager \textbf{Brendan} Rodgers is relishing the occasion.} \\ \hline			
					
					\begin{tabular}{c}Sunday's League (0\%) \\ Hampden when they (50\%) \\ exploit the (80\%)\end{tabular} & \parbox{15.5cm}{Celtic manager Brendan Rodgers says \textbf{Sunday's League} Cup semi-final against Rangers at \textbf{Hampden when they} meet will be a chance to ``\textbf{exploit the} big pitch''.} \\ \hline
					
					\begin{tabular}{c}Sunday's League (40\%) \\Hampden when they (30\%) \\ exploit the (10\%)\end{tabular} & \parbox{15.5cm}{Celtic will look to \textbf{exploit the} atmosphere at \textbf{Hampden when they} face Rangers in \textbf{Sunday's League} Cup semi-final, says manager Brendan Rodgers.} \\ \hline
					
					Keyword \& Position  & Generated text (Error cases) \\ \hline
					
					\begin{tabular}{c}Manager Brendan (80\%) \\meet (50\%) \end{tabular} & \parbox{15.5cm}{Brendan Rodgers is relishing his first visit to Hampden Park as Celtic manager when his side meet Rangers in the Scottish Cup semi-final on Wednesday, 7 May.} \\ \hline
					
					\begin{tabular}{c}semi-final. (40\%) \end{tabular} & \parbox{15.5cm}{New Celtic manager Brendan Rodgers is relishing the prospect of making his debut at Hampden Park in Saturday's Scottish Cup semi-final.} \\ \hline

				\end{tabular}
			}
		}
		\caption{Examples of generated texts from the XSum dataset. In the second error case, ``semi-final.'' (the last word is a period) is given as a keyword. In such case, when the position where the keyword should be placed is obvious, the model tends to ignore the positional control token and place the keyword in the appropriate position.}
		\label{tab:keyword_position_example_appendix2}
	\end{table*}
	
	\begin{table*}[!t]
		\centering
		
		\scalebox{0.80}{
			{\tabcolsep=2.0pt
				\begin{tabular}{c|c} \hline
					
					\multicolumn{2}{c}{Gold text} \\ \hline
					\multicolumn{2}{c}{\parbox{19.5cm}{Bob wanted a car of his own to drive to work. Bob began saving money from his paycheck to buy a car. Eventually Bob saved enough money to buy his very own car. Bob went to a local car dealership and browsed the inventory. Bob found a car he liked and bought it from the dealership.}} \\ \hline
					
					Keyword \& Position  & Generated text \\ \hline

					\begin{tabular}{c} dealership \\ (w/o Position Control)\end{tabular} & \parbox{15.5cm}{I went to the \textbf{dealership} to buy a new car. I was very nervous because I didn't know what I wanted. I went to the dealer and asked them what I wanted. They told me that I wanted a convertible with a leg drive. I drove the car home and it was a great car.} \\ \hline
					
					dealership (0\%) & \parbox{15.5cm}{I went to the \textbf{dealership} to look at a car. I was very nervous about buying a car. I asked the salesman if I could try out a certain car. The salesman told me that I could try out a blue car. I drove the blue car home and loved it so much, I bought it.} \\ \hline
					
					dealership (30\%) & \parbox{15.5cm}{I went to the \textbf{dealership} to buy a new car. I was very nervous about the car and the price. I went to the \textbf{dealership} and looked at the price. I decided to buy the car and I was very happy with it. I am glad I went to the \textbf{dealership} because it was a great deal.} \\ \hline
					
					dealership (80\%) & \parbox{15.5cm}{I went to the \textbf{dealership} to look at a car. I was very nervous because I didn't know what I wanted. I asked the salesman if I could pick out a car. The salesman told me that I could get a new car if I paid \$40,000. I drove to the \textbf{dealership} and bought a new car.} \\ \hline				
					
					saving money (10\%) & \parbox{15.5cm}{I was \textbf{saving money} for a new car. I had been \textbf{saving money} for a long time and was ready to buy it. I went to the car dealership and looked at all the models. I decided on a Ford Mustang and immediately signed up. I drove it home and was so happy with my purchase.} \\ \hline
					
					saving money (60\%) & \parbox{15.5cm}{I was in a hurry to get to work. I had to hurry because I had to hurry so fast. I was going fifteen miles an hour faster than I was. I was \textbf{saving money} so I could get to work. I was glad I was able to make it to work on time.} \\ \hline
					
					saving money (80\%) & \parbox{15.5cm}{I was in a hurry to get to work. I had to hurry because I didn't have my car keys. I looked everywhere for my keys, but couldn't find them. I finally found them under my coat hanger, in my coat pocket. I was relieved that I had \textbf{saved money} on my car keys.''} \\ \hline

					\begin{tabular}{c}drive to work (50\%) \\ saved enough (70\%)\end{tabular} & \parbox{15.5cm}{I was driving to work one day when I saw a car in the road. I pulled over and asked if I could \textbf{drive to work}. The driver told me that he had \textbf{saved enough} money to buy a new car. I drove to work and paid him back. I drove to work and paid him back and he was very happy.} \\ \hline
					
					\begin{tabular}{c}drive to work (20\%) \\ saved enough (0\%)\end{tabular} & \parbox{15.5cm}{I \textbf{saved enough} money to buy a new car. I went to the car dealership to test drive my new car. I drove the car for a few hours before I left. When I got home, I realized I had forgotten my wallet. I had to \textbf{drive to work} to get my wallet back, but I was happy.} \\ \hline
					
					Keyword \& Position  & Generated text (Error cases) \\ \hline
					
					\begin{tabular}{c}began saving money (90\%) \end{tabular} & \parbox{15.5cm}{I was in a hurry to get to work. I had to hurry because I didn't have my car keys. I looked everywhere for my keys, but couldn't find them. I finally found them under my coat, and I was relieved. I was able to grab my keys and walk to work without losing my car keys.} \\ \hline
					
					\begin{tabular}{c}local (0\%) \\ Bob (40\%) \\ enough (60\%) \end{tabular} & \parbox{15.5cm}{\textbf{Bob} was a \textbf{local} handyman. He was hired to fix up a leaky roof on his home. \textbf{Bob} was very handy and did a good job at it. Unfortunately, the roof was too deep and the water would not come out. \textbf{Bob} had to call a \textbf{local} handyman to fix the leaky roof.} \\ \hline
					
				\end{tabular}
			}
		}
		\caption{Examples of generated texts from the ROCStories dataset. Examples of generated texts from the CNN/DM dataset. This table is the complete version of Table~\ref{tab:keyword_position_example2} with gold story and additional examples. As can be seen, even if the keyword ``saving money'' is given to the model, for example, it may be used as another expression, such as ``saved money''. Note that in the quantitative evaluation of this study, such cases where the keyword was paraphrased are classified as ``the keyword was not included in the text''. In some cases, the given keyword is included in the text more than once.}
		\label{tab:keyword_position_example_appendix3}
	\end{table*}
	
\end{document}